\documentclass[10pt,twocolumn,letterpaper]{article}

\usepackage[pagenumbers]{iccv} 

\definecolor{iccvblue}{rgb}{0.21,0.49,0.74}
\usepackage[pagebackref,breaklinks,colorlinks,allcolors=iccvblue]{hyperref}
\usepackage{tikz}
\usepackage{makecell} 

\usepackage{algorithm}
\usepackage[noend]{algpseudocode}
\usepackage{caption}
\usepackage{setspace}       
\usepackage{pifont}
\usepackage{amsmath,amsfonts,bm}
\usepackage{multirow}
\usepackage[accsupp]{axessibility}

\newcommand{\rf}{vanilla RF model}
\newcommand{\oursfull}{Boundary-enforced Rectified Flow Model}
\newcommand{\ours}{Boundary RF Model}
\newcommand{\oursdualpass}{Subtraction-based Boundary RF Model}
\newcommand{\oursmask}{Mask-based Boundary RF Model}

\newcommand{\xmark}{\ding{55}}%

\def\1{\bm{1}}
\def\vv{{\bm{v}}}
\renewcommand{\vec}[1]{{\boldsymbol{#1}}}

\newcommand{\E}{\mathbb{E}}

\newcommand{\X}{\vec{X}}
\newcommand{\Z}{\vec{Z}}

\newcommand{\R}{\mathbb{R}}

\newcommand{\mat}[1]{\boldsymbol{#1}}
\newcommand{\xxi}{\boldsymbol{\xi}}

\newcommand{\atphantom}{\vphantom{${}^2$}}
\newcommand{\AProcedure}[2]{\Procedure{\smash{#1}}{\smash{#2}}}

\newcommand{\AState}[1]{\State{\smash{#1}}}
\newcommand{\AFor}[1]{\For{\smash{#1}}}

\renewcommand{\paragraph}[1]{%
  \vspace{5pt}%
  \noindent%
  \textbf{#1} %
}

\def\paperID{2504} 
\def\confName{ICCV}
\def\confYear{2025}

\title{Improving Rectified Flow with Boundary Conditions}

\author{Xixi Hu$^{1,2}$\thanks{Equal contribution}, Runlong Liao$^{1}$\footnotemark[1], Keyang Xu$^{2}$, \\
Bo Liu$^{1}$, Yeqing Li$^{2}$, Eugene Ie$^{2}$, Hongliang Fei$^{2}$ and Qiang Liu$^{1}$\\
$^1$University of Texas at Austin, $^2$Google\\
{\tt\small \{hxixi,liaorl,bliu,lqiang\}@cs.utexas.edu} \\
{\tt\small \{keyangxu,yeqing,eugeneie,hongliangfei\}@google.com}
}

\begin{document}
\maketitle
\begin{abstract}
Rectified Flow offers a simple and effective approach to high-quality generative modeling by learning a velocity field. However, we identify a limitation in directly modeling the velocity with an unconstrained neural network: the learned velocity often fails to satisfy certain boundary conditions, leading to inaccurate velocity field estimations that deviate from the desired ODE. This issue is particularly critical during stochastic sampling at inference, as the score function's errors are amplified near the boundary. To mitigate this, we propose a \oursfull{}~(\ours{}), in which we enforce boundary conditions with a minimal code modification. \ours{} improves performance over \rf{}, demonstrating 8.01\% improvement in FID score on ImageNet using ODE sampling and 8.98\% improvement using SDE sampling.
\end{abstract}   
\vspace{-5pt}
\section{Introduction}
\label{sec::introduction}
\vspace{-5pt}

Diffusion models have achieved significant success as a powerful class of generative models, demonstrating impressive results across various domains~\citep{song2019generative, song2020score, ho2020denoising}.  More recently, flow-based generative models have emerged as a compelling alternative for generative modeling~\citep{liu2022flow, lipman2022flow, albergo2022building, heitz2023iterative}. Within this family of flow-based models, Rectified Flow (RF)~\cite{liu2022flow, liu2022rectified}, also known as Flow Matching~\cite{lipman2022flow, albergo2022building} stands out by formulating generative modeling as solving an Ordinary Differential Equation (ODE) that transforms a noise distribution into the target data distribution.

The core of RF lies in learning a velocity field that governs this continuous transformation. This is achieved by modeling a linear interpolation between the noise and data distributions along the defined trajectory.  During inference, novel data samples are synthesized through numerical integration of this learned ODE, taking advantage of the powerful generalization capabilities of deep neural networks in high-dimensional spaces. Recent advances have showcased the remarkable scalability of RF in complex generation tasks, including text-to-image and video generation, yielding compelling results~\cite{esser2024scaling, blackforestlabs_flux_2023, polyak2024movie}. 
Furthermore, the inherent property of RF to encourage straighter trajectories enables efficient numerical discretization, such as with Euler samplers, leading to accelerated sampling speeds – a crucial advantage for practical deployment.

\begin{figure}[t!]
    \centering
    \vspace{-5pt}
    \includegraphics[width=\columnwidth]{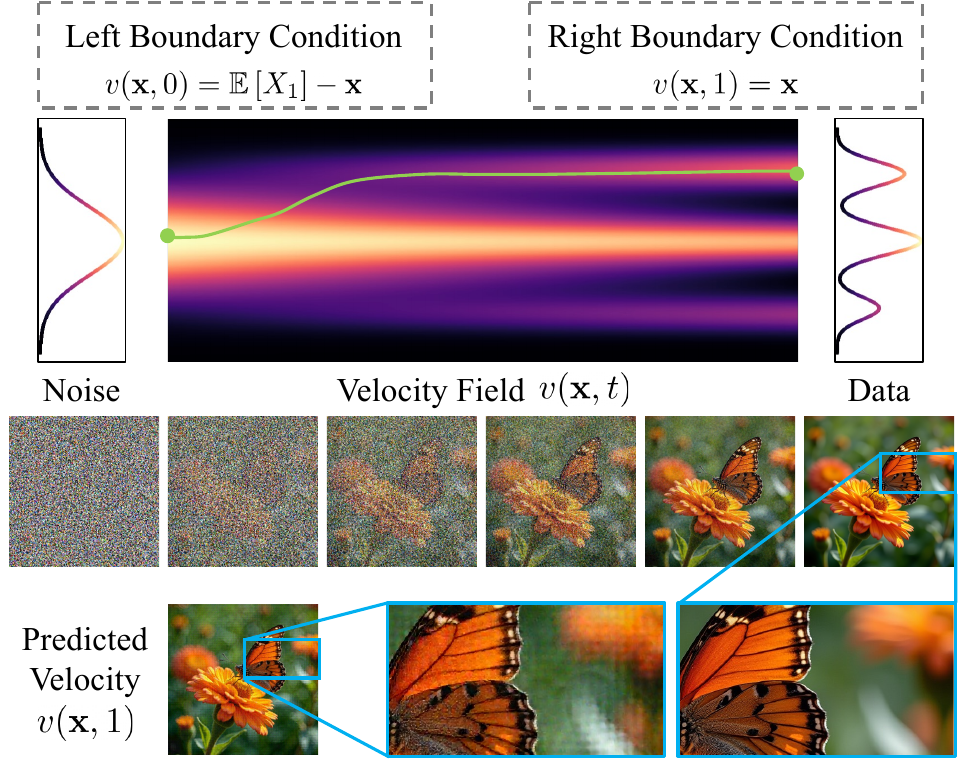}
    \caption{\textbf{Boundary condition violation in Rectified Flow: predicted vs. expected velocity.}  Vanilla Rectified Flow model learns a velocity field $v(\textbf{x}, t)$ to transform noise (left, $t=0$) into data (right, $t=1$). Ideally, this learned velocity field should satisfy defined boundary condition (top). However, as visualized, the predicted velocity at $t=1$ (bottom) deviates from the expected data distribution and violate the right boundary condition $v(\textbf{x}, t) = \textbf{x}$. This highlights a critical practical limitation of \rf. Please refer to Appendix~\ref{sec::flux} for more examples. }
    \label{fig:teaser}
    \vspace{-5pt}
    \vspace{-5pt}
\end{figure}

Despite these advances, 
we observe a critical limitation when training RF models in practice: the learned velocity field often exhibits suboptimal behavior, particularly in violating a set of theoretically derived boundaries
as shown in Figure~\ref{fig:teaser}~(upper panel). 
For example, it is theoretically expected that the velocity field reduces to the identity map when applied to the clean images ($t=1$), as shown
Figure~\ref{fig:teaser} (lower panel). 
Notably, this challenge persists even in large-scale text-to-image applications, where \rf{} often struggles to accurately satisfy the theoretical boundary conditions as we demonstrated in Appendix~\ref{sec::flux}. 

This inaccurate velocity field estimation is particularly problematic in stochastic RF sampling, where the added Langevin dynamics term, containing a score estimation component, can diverge near the terminal time ($t=1$), resulting in over-smoothed or cartoonish samples~\cite{hu2024amo}. 

To address these limitations, we propose \emph{\oursfull{}} (\ours{}), a novel approach designed to explicitly enforce the boundary velocities by incorporating constraints and informed parameterizations. Our method effectively rectifies the terminal behavior of the velocity field, ensuring adherence to boundary conditions, while preserving the representational power of state-of-the-art architectures like U-Net and Diffusion Transformers~\cite{peebles2023scalable}. Moreover, \ours{} provides a simple and readily implementable parameterization that enforces boundary conditions, easily integrable with existing Rectified Flow models. 

To validate the efficacy of \ours{}, we conduct a comprehensive ablation study on image generation tasks. Our empirical evaluations demonstrate that \ours{} and its variants, \oursdualpass{}, consistently enhance both deterministic and stochastic sampling performance. For instance, in ImageNet experiment, \ours{} achieves a Fr\'echet Inception Distance (FID) of 6.32, outperforming the Rectified Flow baseline which yields an FID of 6.87. Moreover, by explicitly imposing the boundary conditions, \ours{} also contribute to a more stable score function near $t=1$, further contributing to performance gains of stochastic samplers. In summary, \ours{} offers three primary advantages:

\begin{itemize}

    \item \textbf{Easy to Implement and Adapt:} \ours{} is straightforward to implement and integrates easily into existing Rectified Flow models, offering both mask-based and subtraction-based model variants for flexibility.
    \vspace{3pt}
    \item \textbf{Improved Generation Quality:} \ours{} consistently enhances image generation quality compared to \rf{}.
    \vspace{3pt}
    \item \textbf{Stable Stochastic Sampling:} By enforcing boundary conditions, \ours{} stabilizes the score function, leading to improved and more robust stochastic sampling results. 

\end{itemize}

\vspace{-5pt}
\section{Background}
\label{sec::background}
\vspace{-5pt}

This section provides a concise introduction to Rectified Flow
(RF)~\citep{liu2022rectified}. RF aims to construct an efficient transport mapping from a simple noise distribution \(\pi_0\) to a complex data distribution \(\pi_1\). This framework is related to, but conceptually simpler than, diffusion-based or score-based approaches~\citep{song2019generative, song2020score, ho2020denoising}, as it learns an ODE with a velocity field that matches a prescribed interpolation slope.

Formally, let \(\X_0 \sim \pi_0\) denote noise samples (e.g., from a standard Gaussian) and \(\X_1 \sim \pi_1\) denote data samples. Suppose we are given a coupling \((\X_0, \X_1)\) drawn independently from their respective distributions. Rectified Flow first specifies an interpolation process $\X_t = \mathtt{I}_t(\X_0, \X_1), \, t \in [0,1],$ that smoothly connects \(\X_0\) and \(\X_1\). A common choice is the straight-line interpolation \(\X_t = t\X_1 + (1-t)\X_0\) \citep{liu2022flow, lipman2022flow, albergo2022building, heitz2023iterative}, although different affine interpolations introduce essentially equivalent RF \citep{shaul2023bespoke, gao2025diffusionmeetsflow, let2025Liu}. Note that we cannot directly generate data samples from the interpolation process itself, because sampling \(\X_t\) depends on knowing \(\X_1\). To make the procedure causal, RF then $\mathtt{Rectify}$ this interpolation process into an ODE process of the form
\begin{equation}
\mathrm{d}\Z_t \;=\; \boldsymbol{v}_t(\Z_t)\,\mathrm{d}t,
\quad 
\text{starting from } \Z_0 \sim \pi_0,
\label{equ:odemodel}
\end{equation}
by learning a velocity field \(\vv(\mathbf{x},t)\colon \R^d \times [0,1] \mapsto \R^d\). The ideal velocity \(\vv_t^*\) at each point \(\mathbf{x}\) and time \(t\) is the conditional expectation of the interpolation slope:
\begin{equation}
\boldsymbol{v}_t^*(\mathbf x) =\mathbb{E}\bigl[\dot{\X}_t \,\big|\; \X_t = \mathbf x\bigr],
\end{equation}
where \(\dot{\X}_t\) denotes the derivative \(\mathrm{d}\X_t/\mathrm{d}t\). In practice, \(\vv\) is parameterized by a neural network and is learned by minimizing the mean-squared error between \(\vv(\X_t,t)\) and the slope of \({\X}_t\):
\begin{equation} \label{eq:rf-objective}
\min_{\vv} \int_0^1 \mathbb{E}_{(\X_0, \X_1)\sim \pi_0 \times \pi_1} 
\left[\eta_t \| \vv(\X_t, t) - \dot{\X}_t \|^2 \right] \,\mathrm{d}t
\end{equation}
where \(\eta_t\) is a time-dependent weighting factor. Upon convergence, the learned ODE trajectories \(\{\Z_t\}\) share the same marginal distribution as \(\{\X_t\}\) at each time \(t\) \citep{liu2022flow}. Consequently, \(\Z_1\) should match the target distribution \(\pi_1\). In practice, starting from \(\Z_0 = \X_0\), we can numerically integrate this ODE forward to obtain \(\Z_1\), which serves as generated samples from \(\pi_1\). 

In practice, the objective function in Equation~\eqref{eq:rf-objective} is typically optimized using stochastic gradient descent.  A common strategy is to sample time \(t\) from a uniform distribution over \([0, 1]\) or employ a re-weighted sampling scheme such as the logit-normal distribution~\cite{esser2024scaling}.  With these approaches, time points are sampled independently for each training batch, and the expectation in Equation~\eqref{eq:rf-objective} is approximated by averaging over these sampled time points and data pairs \((\X_0, \X_1)\).  However, such uniform or re-weighted sampling strategies may not explicitly emphasize the boundaries of the flow, i.e., \(t \approx 0\) and \(t \approx 1\).  Consequently, the learned velocity field might be less accurate at these crucial boundary regions, potentially contributing to the observed challenges in satisfying the theoretical boundary conditions in practical RF models.

\paragraph{Euler Sampler}
A standard way to solve RF ODE \eqref{equ:odemodel} is via the explicit Euler method. Let \(0 = t_0 < t_1 < \dots < t_N = 1\) be $N$ timesteps in the interval \([0,1]\).  The discretized trajectory \(\{\tilde{\Z}_{t_k}\}\) evolves according to\[
\tilde{\Z}_{t_{k+1}} 
= 
\tilde{\Z}_{t_k} + \bigl(t_{k+1} - t_k\bigr)\,\vv\bigl(\tilde{\Z}_{t_k}, t_k\bigr),
\quad 
\tilde{\Z}_0 \sim \pi_0,
\]
\paragraph{Stochastic Samplers}
When an ODE is solved numerically (e.g., via Euler), both model approximation and discrete-time updates can introduce errors. These errors accumulate over time, causing the estimated distribution to drift from its true counterpart. To mitigate this drift, one can add a feedback mechanism in the form of Langevin dynamics to the pretrained ODE model \(\vv(\mathbf{x},t)\)~\cite{song2020score, let2025Liu, hu2024amo}:
\begin{align}
\small
\mathrm{d}\Z_t = 
\underbrace{\vv\bigl(\Z_t, t\bigr)\,\mathrm{d}t}_{\textcolor{blue}{\text{Flow}}}
+
\underbrace{\sigma_t\,\nabla \log \rho_t\bigl(\Z_t\bigr)\,\mathrm{d}t 
+ \sqrt{2\sigma_t}\,\mathrm{d} \boldsymbol{W}_t}_{\textcolor{magenta}{\text{Langevin}}}, \label{equ:sde-sampler}
\end{align}
where $\rho_t$ is the density function of $\Z_t$ following the ODE model \eqref{equ:odemodel}, $\sigma_t$ is a non-negative sequence, and $\boldsymbol{W}_t$ is a standard Brownian motion. This Langevin force nudges \(\Z_t\) toward its own distribution \(\rho_t\) but does not alter the marginal. As a result, \(\Z_t\) under this SDE retains the same marginal distribution as the original ODE~\eqref{equ:odemodel}.

In the standard RF setting, when $\pi_0 \sim \mathcal{N}(\mathbf{0}, \mat{I})$ and $\X_0 \perp \X_1$, the score function \(\nabla \log \rho_t\) can be estimated from \(\vv_t\) using Tweedie's formula:
\begin{equation}
\label{equ:tweedie}
\nabla \log \rho_t(\mathbf{x}) = \frac{t\vv(\mathbf{x}, t) - \mathbf{x}}{1 - t}.
\end{equation}
Note, however, that if \(\vv(\mathbf{x},t)\) is not accurately learned, relying on a relatively large \(\sigma_t\) can amplify the estimation errors for \(\nabla \log \rho_t(\mathbf{x})\). In text-to-image generation tasks, choosing a relatively large \(\sigma_t\) often manifests as oversmoothing and loss of fine details in synthesized images~\citep{hu2024amo}.

\vspace{-5pt}
\section{Boundary-enforced Rectified Flow Models}
\label{sec::method}
\vspace{-5pt}

In this section, we first analyze the boundary conditions under linear interpolation, revealing the inherent problem of unconstrained velocity fields. We then introduce two variants of \ours{}: 1) \oursmask{} that explicitly enforces boundary constraints into the velocity parameterization, and 2) \oursdualpass{}, which offers design simplicity and strong empirical performance. We further discuss the advantage of \ours{} in estimating the score function and stochastic sampling in generative models. 

\subsection{Boundary Conditions of Rectified Flow}
As detailed in Section \ref{sec::background}, the optimal velocity field for Rectified Flow, \(\X_t = (1 - t)\,\X_0 + t\,\X_1\), is given by:
\begin{equation}
\vv^*(\mathbf{x}, t) = \E_{\X_0, \X_1} \left[ \X_1 - \X_0 \mid \X_t = \mathbf{x} \right].
\end{equation}
Given the assumption that the noise $\X_0$ and data $\X_1$ are independent variables, the boundary velocities at \(t=0\) and \(t=1\) are: 
\begin{align*}
    \vv^*(\mathbf{x}, 0) &= \E \left[ \X_1 -\X_0 \mid \X_0 = \mathbf{x} \right] = \E \left[ \X_1\right] - \mathbf{x}, \\
    \vv^*(\mathbf{x}, 1) &= \E \left[ \X_1 -\X_0 \mid \X_1 = \mathbf{x} \right] = \mathbf{x} - \E \left[ \X_0\right] = \mathbf{x},
\end{align*}
where the simplification at \(t=1\) leverages the standard RF setting \(\X_0 \sim \mathcal{N}(\mathbf{0}, \mat{I})\), resulting in \(\mathbb{E}[\X_0] = \mathbf{0}\).

However, in practice, a general neural network \(\vv_\theta(\mathbf{x},t)\) is used to approximate \(\vv^*(\mathbf{x},t)\), and this approximation often fails to adhere to the derived boundary conditions, as illustrated in Figure~\ref{fig:teaser}.

\subsection{Mask-based Boundary RF Model} 

To ensure adherence to the boundary conditions of Rectified Flow, we propose a boundary-aware parameterization of the velocity field, \(\vv(\mathbf{x}, t)\). This parameterization enforces both boundary conditions: \(\vv(\mathbf{x},1) = \mathbf{x}\) and \(\vv(\mathbf{x},0) = C - \mathbf{x}\), where \(C = \mathbb{E}[\X_1]\).
We achieve this by structuring the velocity field as follows:
\begin{equation}~\label{eq:our-v}
\vv(\mathbf{x}, t) = g(t) \cdot (C - \mathbf{x}) + f(t) \cdot \mathbf{x} + h(t) \cdot m_\theta(\mathbf{x}, t),
\end{equation}
where \(m_{\theta}(\mathbf{x},t): \mathbb{R}^d \times \mathbb{R} \to \mathbb{R}^d\) is a neural network (e.g., U-Net or DiT), and \(f\), \(g\), and \(h: \mathbb{R} \to \mathbb{R}\) are scalar functions with the following properties:
\begin{align*} 
& g(1) = f(0) = h(0) = h(1) = 0, \\
& g(0) = f(1) = 1.
\end{align*}
\noindent
This construction guarantees that both boundary conditions are satisfied by design. While various functions can fulfill these conditions (see Section~\ref{sec::ablation} for examples), a natural and effective choice is: 
\begin{equation} 
g(t) = \cos\bigr(\frac{\pi}{2}t\bigr),~~~
f(t) = \sin\bigr(\frac{\pi}{2}t\bigr),~~~
h(t) = \sin(\pi t).
\end{equation}
This results in the loss function of Eq.~\eqref{eq:rf-objective} in \rf{} turning into
\begin{equation} \label{eq:ours-rf-objective}
\min_{\theta} \int_0^1 \mathbb{E}_{(\X_0, \X_1)\sim \pi_0 \times \pi_1}
\left[\eta_t || \vv(\X_t, t) - \dot{\X}_t ||^2 \right] \mathrm{d}t,
\end{equation}
\noindent
where \(v(\X_t, t)\) is now parameterized by our boundary-aware velocity field from Eq.~\eqref{eq:our-v}.

We further explore alternative function choices and their empirical performance in Section~\ref{sec::ablation}.

\subsection{Subtraction-based Boundary RF Model}
While our \oursmask{} effectively enforces both \(t=1\) and \(t=0\) boundary conditions, in practice, we found that only ensuring the \(t=1\) condition (\(\vv(\mathbf{x},1)=\mathbf{x}\)) also empirically benefits high-quality generation. Motivated by this observation and the desire for a less constrained parameterization, we introduce a subtraction-based variant:
\begin{equation}
\vv(\mathbf{x},t)=x + m_{\theta}(\mathbf{x},t) - m_{\theta}(\mathbf{x},1).
\end{equation}
This construction inherently satisfies \(\vv(\mathbf{x},1) =\mathbf{x}\) by design: \(\vv(\mathbf{x},1) = \mathbf{x} + m_{\theta}(\mathbf{x},1) - m_{\theta}(\mathbf{x},1) = \mathbf{x}\), regardless of the neural network \(m_{\theta}\).  This guarantee is achieved without carefully tuning scalar functions \(f, g, h\), offering a more direct and flexible approach.

Compared to the double-boundary parameterization, the \oursdualpass{} provides design simplicity by directly leveraging a standard neural network output.  However, this simplicity introduces a computational trade-off: Each evaluation of \(\vv(\mathbf{x},t)\) requires computing \(m_{\theta}(\mathbf{x},1)\) in an additional forward pass. We will empirically compare the performance and characteristics of both the \oursmask{} and \oursdualpass{} in Section~\ref{sec::ablation}.

\begin{figure}[ht!]
    \centering
    \includegraphics[width=1.\columnwidth]{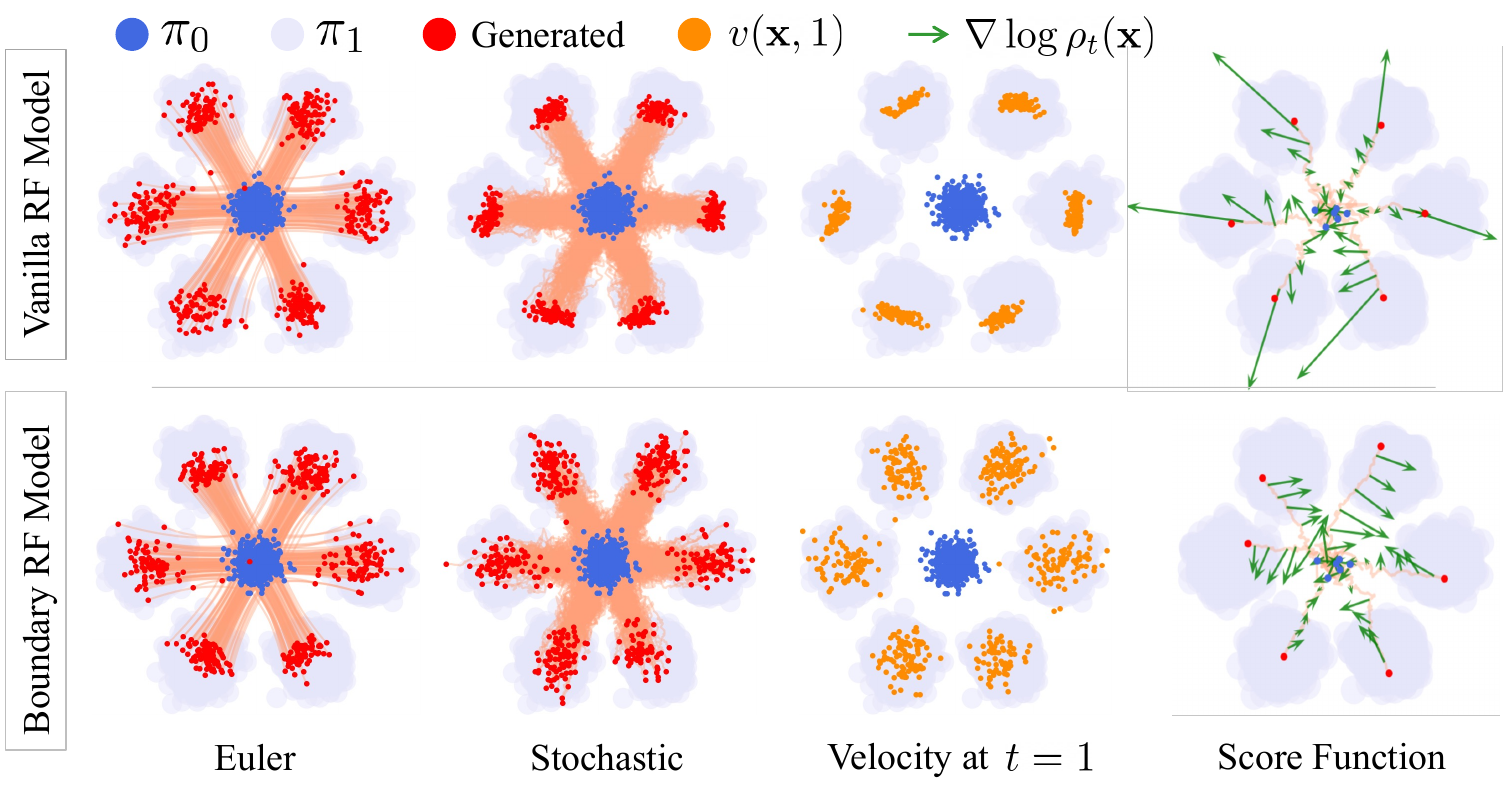}
    \caption{\textbf{Toy example: \ours{} stabilizes stochastic sampling and score function.} We visualize the behavior of \rf{} and \ours{} when learning to map from noise \textcolor{blue}{$\pi_0$} to data \textcolor{red}{$\pi_1$}. From left to right, we visualize the followings: 1) Euler Sampling (Deterministic): Both models learn effective ODE trajectories, generating similar samples via deterministic Euler sampler. 2) Stochastic Sampling: \rf{} produces more concentrated samples due to score function instability. In contrast, \ours{}, by enforcing boundary conditions, generates samples that retain the shape of the target distribution. 3) Velocity at $t=1$ Boundary: \rf{} velocities deviate from the data, violating \(\vv(\mathbf{x}, 1) = \mathbf{x}\), while \ours{} adheres to this boundary condition. 4) Score Function: We visualize the score function computed using Eq.~\eqref{equ:tweedie}. Visualization of the score function near \(t=1\) shows that \rf{} exhibits an unstable and divergent score field. In contrast, \ours{} demonstrates a stable and well-behaved score function, preventing the unboundedness that leads to concentration effect in stochastic sampling.}~\label{fig:sde_toy}
    \vspace{-5pt}
    \vspace{-5pt}
\end{figure}

\subsection{Impact on Stochastic Sampling}

As discussed, stochastic samplers for Rectified Flow inherently rely on the score function, which, via Tweedie's formula (Eq.~\eqref{equ:tweedie}), is inversely proportional to \((1-t)\).  Consequently, if the boundary condition \(\vv(\mathbf{x},1)=x\) is not enforced, the score function estimate will become unbounded as \(t \to 1\). This unboundedness destabilizes stochastic sampling, particularly when using larger step sizes (or \(\sigma_t\)) for faster generation. Such instability can manifest as over-smoothing and loss of detail in generated samples~\cite{hu2024amo}. In contrast, \ours{}, by enforcing \(\vv(\mathbf{x},1)=\mathbf{x}\), prevents the score function from diverging. This leads to a more stable score estimate, enabling robust stochastic sampling with larger \(\sigma_t\). 

To visually illustrate this stabilization, we present a toy example in Figure~\ref{fig:sde_toy}, directly comparing constrained and unconstrained models under stochastic sampling and demonstrating the improved score function behavior. We further showcase the benefits of \ours{} in the more complex domain of image generation. As shown in Section~\ref{sec::impact_stochastic_sampling}, \ours{} consistently produces sharper and more detailed image samples compared to \rf{} when using stochastic samplers, effectively alleviating over-smoothing artifacts.

\section{Related Work}
\label{sec:related}

\paragraph{Generative modeling with diffusion and flow models.}
Diffusion models~\cite{ho2020denoising, song2020score, song2019generative} have revolutionized the field of generative modeling by introducing iterative refinement for content generation. These models have demonstrated the capacity to generate samples from complex distributions across various data modalities, including images, videos, audio, and point clouds~\cite{dhariwal2021diffusion, saharia2022photorealistic, ho2022video, podell2023sdxl, betker2023improving, baldridge2024imagen}.  A key characteristic of diffusion models is their reliance on iterative sampling over numerous steps, with the continuous limit being described by a Stochastic Differential Equation (SDE) process. In contrast, Rectified Flow ~\cite{liu2022flow, lipman2022flow, albergo2022building, heitz2023iterative}, also known as Flow Matching, InterFlow, or IADB, presents an alternative approach to generative modeling based on Ordinary Differential Equations (ODEs). This ODE-based formulation is conceptually simpler and has proven to be successful in diverse practical applications, achieving performance comparable to or exceeding that of diffusion models in many cases~\cite{esser2024scaling, fei2024flux, le2024voicebox, polyak2024movie}.  Recent research has increasingly recognized the conceptual connections and transferability between diffusion models and flow-based models~\cite{let2025Liu, gao2025diffusionmeetsflow, lipman2024flow, kingma2023understanding}. Consequently, sampling approaches, training techniques, and other methodological details are being explored for interchangeable use between these frameworks~\cite{hu2024amo, let2025Liu}. In this work, we primarily focus on Rectified Flow due to its current efficacy and practical advantages in various applications. However, we note that the core concepts and contributions of our approach could also be potentially extended to diffusion models with minor adaptations.

\paragraph{Training enhancements for diffusion and flow models.}
Recent advances have improved flow and diffusion models' training by adjusting the interpolation path or the noise schedule~\cite{nichol2021improved,lee2024improving,kingma2023understanding}, or by explicitly re-weighting the loss across noise levels~\cite{karras2022elucidating,esser2024scaling,kingma2023understanding}. A series of analyses~\cite{let2025Liu,gao2025diffusionmeetsflow,lipman2024flow,kingma2023understanding} reveals that the noise schedule itself adds no expressiveness: different choices merely impose alternative weightings over noise levels. For instance, EDM~\cite{karras2022elucidating} decouples the noise schedule from the auxiliary time variable, enabling sampling at arbitrary noise scales during inference. While other works~\cite{kingma2023understanding,let2025Liu,gao2025diffusionmeetsflow} prove that any monotonic noise schedule reduces to a log-SNR-weighted loss. These methods, however, do not focus on the study of boundary conditions on RF. By contrast, our approach explicitly enforces the theoretical boundary conditions of the velocity network, remedying a shortcoming of standard RF parameterizations, offering a distinct perspective from existing training refinements.

\paragraph{Stochastic sampling methods for diffusion and flow models.}
Deterministic sampling methods, often based on ODE solvers like the Euler method, are commonly employed in both diffusion models~\cite{song2020score, song2020denoising, lu2022dpm, karras2022elucidating} and Rectified Flow models~\cite{hu2024amo}.  Stochastic sampling methods offer an alternative by introducing randomness into the sampling process, potentially generating higher quality samples~\cite{meng2023distillation, karras2022elucidating, hu2024amo}. These techniques, often leveraging Langevin dynamics or related stochastic differential equation (SDE) solvers, implicitly benefit from accurate score function estimation. Our \oursfull{} directly enhances the stability of the score function by explicitly enforcing boundary conditions. This improved score function behavior, as we demonstrate, is particularly advantageous for stochastic samplers, enabling them to operate more robustly, especially with larger step sizes for faster sampling.

\vspace{-5pt}
\section{Experiments}
\label{sec::experiments}

In this section, we present comprehensive experiments to empirically validate the effectiveness of \ours{} and address the following key research questions.

\begin{itemize}
    \item \textbf{Efficacy of Boundary Model with Rectified Flow:} Does the integration of the boundary condition improve the performance of Rectified Flow (Section~\ref{sec::rf_comparison})?
    
    \item \textbf{Ablation Study of Boundary Model Variants:} How do different variations of the proposed \ours{} impact the generative performance (Section~\ref{sec::ablation})?

    \item \textbf{Robustness to Stochastic Sampling:} Does \ours{} enhance the stability and quality of stochastic sampling in Rectified Flow (Section~\ref{sec::impact_stochastic_sampling})?
    
    \item \textbf{Scalability and High-Resolution Performance:} How does \ours{} scale to larger models and higher-resolution image generation tasks (Section~\ref{sec:scalling_law})?
\end{itemize}

\subsection{Experimental Setup}

\paragraph{Datasets.} 
We evaluated our model on CIFAR-10~\cite{krizhevsky2009learning} ($32 \times 32$ resolution) and ImageNet~\cite{deng2009imagenet} ($256 \times 256$ and $512 \times 512$ resolutions), two widely used image datasets.

\paragraph{Evaluation Metrics.}  
To quantitatively assess the generative performance, we employ standard metrics commonly used in image generation literature. These include Fr\'echet Inception Distance (\textbf{FID})~\cite{heusel2017gans}, Inception Score (\textbf{IS})~\cite{salimans2016improved}, and \textbf{Precision and Recall (P/R)}~\cite{sajjadi2018assessing}.

\paragraph{Model Architecture and Training Details.} 
We build upon the Diffusion Transformer (DiT) architecture~\cite{peebles2023scalable} for our generative models.  Unless otherwise specified, the main results are reported using the DiT-B/2 configuration, which comprises 12 layers, a hidden dimension of 768, and 12 attention heads. 
All models were implemented using JAX~\cite{bradbury2018jax} and trained on TPU-v5 pods. To ensure a fair comparison across all experiments, we maintain consistent random seeds and data preprocessing procedures. For stochastic sampling, we primarily employ the Stochastic Curved Euler Sampler~\cite{let2025Liu} throughout our main experiments, which subsequently referred to as the "SDE sampler". Meanwhile, we also explore Overshooting Sampler~\cite{hu2024amo} which offers controllable stochasticity strength. Further details regarding training hyperparameters and implementation specifics are provided in Appendix~\ref{sec::training-details}. 

\subsection{Comparison with Vanilla RF Model}
\label{sec::rf_comparison} 

In this section, we present a comparative analysis of our proposed method, \ours{}, against the \rf{}.  We evaluate both quantitatively and qualitatively, demonstrating the better generative performance of \ours{}.

\begin{table*}[ht!]
\centering
\resizebox{1.0\textwidth}{!}{
    \setlength{\tabcolsep}{0.3cm}
    \begin{tabular}{l|cccc|cccc} 
    \toprule
    & \multicolumn{4}{c|}{\textbf{Euler Sampler}} & \multicolumn{4}{c}{\textbf{SDE Sampler}} \\ 
    & \textbf{FID~$(\downarrow)$} & \textbf{IS~$(\uparrow)$} & \textbf{Precision~$(\uparrow)$} & \textbf{Recall~$(\uparrow)$} & \textbf{FID~$(\downarrow)$} & \textbf{IS~$(\uparrow)$} & \textbf{Precision~$(\uparrow)$} & \textbf{Recall~$(\uparrow)$}\\ 
    \midrule
    \texttt{CIFAR-10} &  &  &  &  &  &  &  & \\
    Vanilla RF Model &3.94  &9.23  &\textbf{0.71}  &0.57  &3.63  &9.61  &\textbf{0.73}  &0.56 \\ 
    \oursmask{} & {3.75}  &9.36  &\textbf{0.71}  &0.57  &3.53  &\textbf{9.67}  &\textbf{0.73}  &0.56 \\
    \oursdualpass{} &\textbf{3.47}  &\textbf{9.37}  &0.70  &\textbf{0.58}  &\textbf{3.04}  &9.56  &0.72  &\textbf{0.57}  \\
    \midrule
    \texttt{ImageNet $256 \times 256$} &  &  &  &  &  &  &  & \\
    Vanilla RF Model & 6.87 &  130.08 &  \textbf{0.67} &  0.58 &  6.79 &  141.86 &  0.67 &  0.57 \\
    RF with Mode Sampling & 7.48 & 114.84 & \textbf{0.67} & 0.57 & 7.56 & 120.97 & 0.67 & 0.56 \\
    \oursmask{}       & 6.63 &  122.35 &  \textbf{0.67} &  \textbf{0.60} &  \textbf{6.18} &  141.43 &  \textbf{0.69} &  0.56 \\
    \oursdualpass{} & \textbf{6.32} &  \textbf{137.28} &  0.66 &  0.59 &  6.38 &  \textbf{149.00} &  0.66 & \textbf{0.59} \\
    \midrule
    \texttt{\texttt{ImageNet $256 \times 256$} with CFG} &  &  &  &  &  &  &  & \\ 
    Vanilla RF Model & 6.11 & 201.64 & 0.74 & 0.52 & 7.20 & 224.36 & 0.75 & 0.50 \\
    RF with Mode Sampling & 5.60 & 185.37 & 0.75 & 0.51 & 6.60 & 197.11 & 0.75 & 0.49 \\
    \oursmask{}        & \textbf{4.94} & 192.14 & \textbf{0.75} & \textbf{0.53} & \textbf{6.52} & 222.46 & \textbf{0.77} & 0.49 \\
    \oursdualpass{}  & 5.68 & \textbf{216.45} & \textbf{0.75} & \textbf{0.53} & 6.94 & \textbf{238.38} & 0.75 & \textbf{0.51} \\
    \bottomrule
    \end{tabular}
    }
    \caption{Quantitative evaluation of \ours{} against the \rf{} across different datasets and metrics.  Lower FID and higher IS, Precision, and Recall indicate better generative performance.} 
    \label{tab:main-results}
    \vspace{-5pt}
    \vspace{-5pt}
\end{table*}

\paragraph{Quantitative Evaluation.}
\label{sec::quantitative_rf_comparison}
Table~\ref{tab:main-results} summarizes the quantitative performance of \ours{} and \rf{} on CIFAR-10 and ImageNet $256 \times 256$. As shown, \ours{} exhibit strong quantitative performance, often outperforming \rf{}, especially in FID across both datasets and sampler types. across both datasets and evaluation metrics, including FID, IS, Precision, and Recall.  On ImageNet $256 \times 256$, \oursmask{} achieves a notable FID of 6.32, surpassing the 6.87 FID obtained by the \rf{}. In addition, we also compare against Rectified Flow employing mode sampling~\cite{esser2024scaling}.  Mode sampling is a timestep sampling strategy designed to concentrate sampling probability towards the boundaries of the timestep range (i.e., near 0 and 1). However, even when compared to Rectified Flow with this enhanced mode sampling strategy, \ours{} maintains its performance advantage, highlighting the effectiveness of our boundary-aware approach in addressing the inherent limitations of \rf{}. Moreover, \oursdualpass{} performs the best among all methods in terms of IS score. 

\begin{figure}[ht!]
    \centering
    \includegraphics[width=1.0\columnwidth]{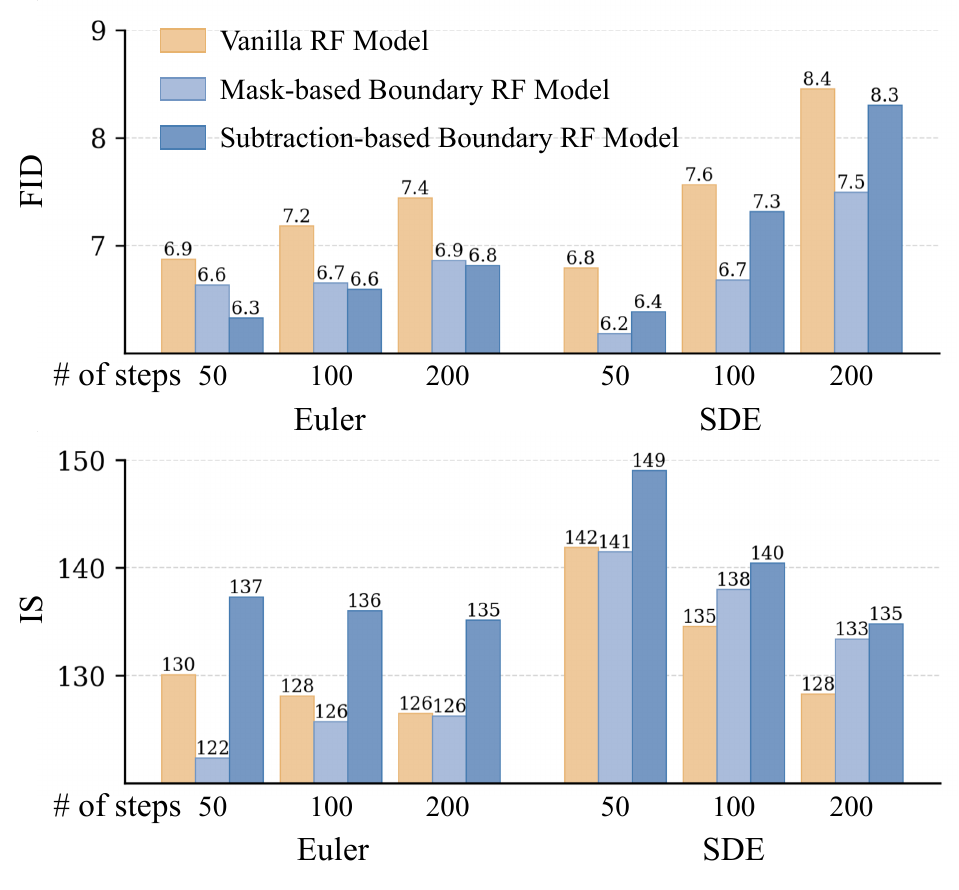}\vspace{-5pt}
    \caption{Performance comparison of \ours{} and \rf{} v.s. sampling steps (50, 100 and 200 steps) on ImageNet dataset. \ours{} consistently outperforms \rf{} across varying numbers of sampling steps, exhibiting a more substantial performance gain at higher step counts. } 
    \label{fig:euler-ours-number}
    \vspace{-5pt}
    \vspace{-5pt}
\end{figure}

Figure~\ref{fig:euler-ours-number} examines the impact of sampling step count on FID and IS scores for \ours{} and \rf{}.  Evaluations are conducted across 50, 100, and 200 steps using both Euler and SDE samplers on ImageNet dataset.  The results demonstrate that \ours{} exhibits better performance at all step counts. While quantitative metrics may indicate a slight performance degradation with increasing sampling steps, this trend may not fully reflect human perceptual quality.  In practice, a higher number of steps often yields images with richer content and finer details, potentially surpassing what standard metrics capture~\cite{esser2024scaling, hu2024amo}. Also note that while the \oursdualpass{} achieves the best performance overall, it requires approximately twice the computation compared to the single-pass \oursmask{} and \rf{}.

\begin{figure*}[ht!]
    \centering
    \includegraphics[width=1.0\textwidth]{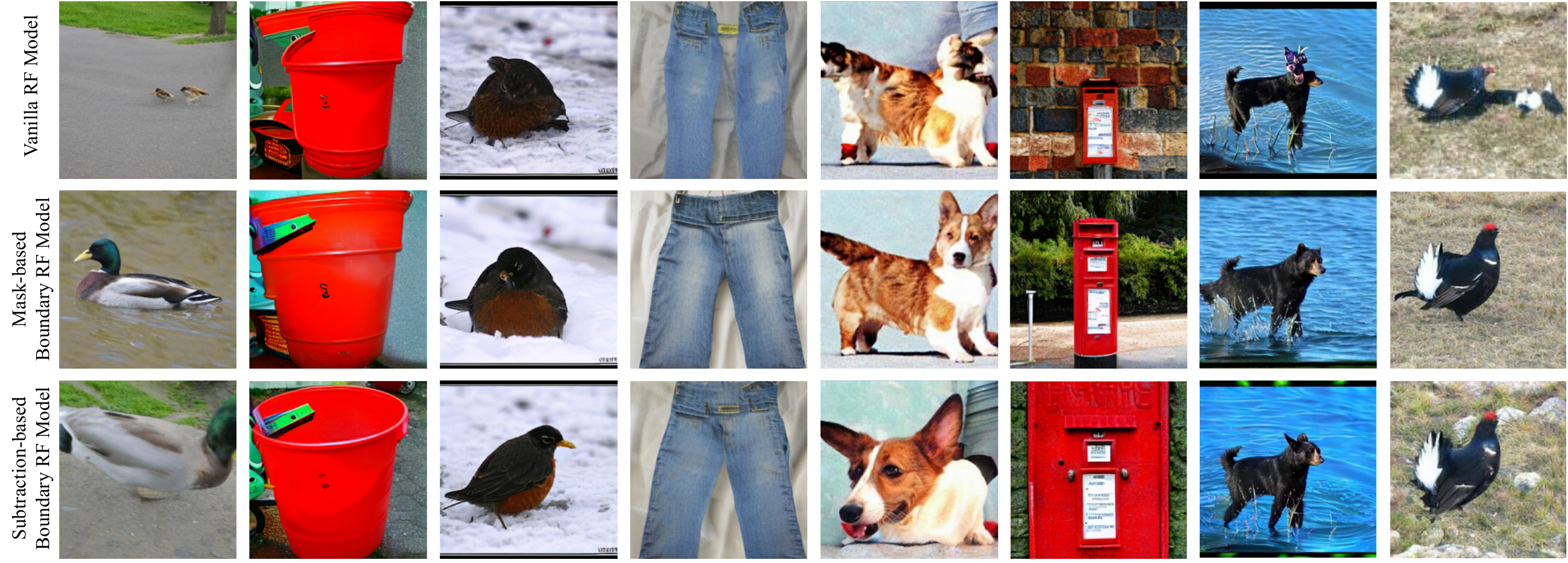}
    \caption{\textbf{Qualitative comparison of image generation results on ImageNet $\mathbf{256 \times 256}$ dataset.} We present paired examples generated by \rf{}, \oursmask{} and \oursdualpass{}. We use the same random seed during training and evaluation for all models to ensure a fair visual comparison.  Our approaches consistently generate images with better structures and improved visual fidelity compared to \rf{}. Additional visual examples are provided in Appendix~\ref{sec::additional-examples}.
    } 
    \label{fig:rf_ours_images}
\end{figure*}

\paragraph{Qualitative Evaluation}
\label{sec::qualitative_rf_comparison} 
Figure~\ref{fig:rf_ours_images} provides a qualitative side-by-side comparison of image generation results from \oursmask{}, \oursdualpass{} and \rf{} on the ImageNet 256×256 dataset. For controlled experimentation, sample generation for all models was conducted using a consistent random seed. Qualitative inspection reveals that images generated by \oursmask{} and \oursdualpass{} exhibit better object fidelity and fewer artifacts. 

\subsection{Ablation Studies}
\label{sec::ablation}

To investigate the contribution of different components within our \ours{} framework, we conduct a detailed ablation study. We primarily focus on evaluating the impact of different design choices in \ours{} velocity parameterization, including the choice of boundary function and the single- versus double-boundary enforcement strategy. all ablation experiments were conducted on ImageNet $256 \times 256$.

\begin{figure}[ht!]
\centering
\includegraphics[width=1.0\columnwidth]{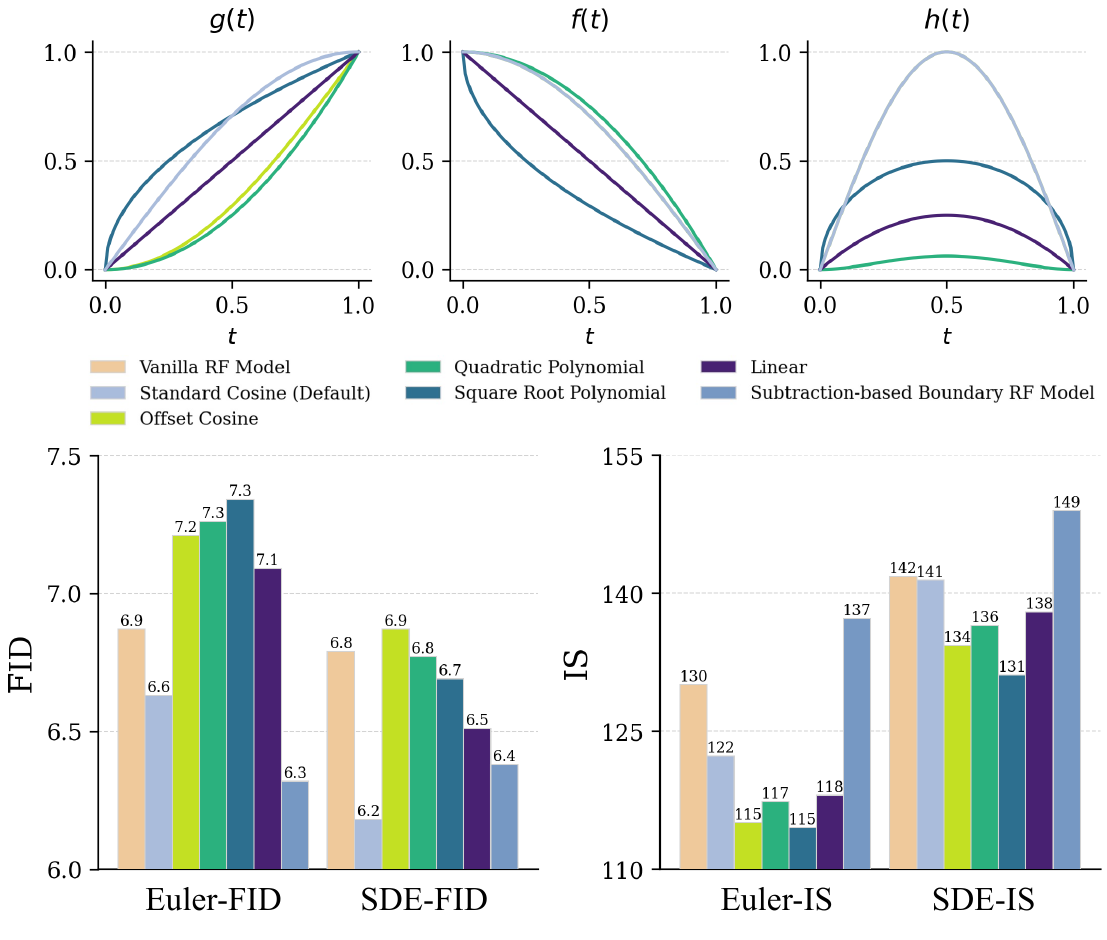}
\caption{Ablation study on boundary functions. Quantitative comparison of different choices for functions (f(t), g(t), h(t)) in the double-boundary model.  Metrics are evaluated on ImageNet $256\times256$ dataset. }
\vspace{-5pt}
\vspace{-5pt}
\label{fig::ablation_functions_numbers}
\end{figure}

\paragraph{Impact of Boundary Functions \texorpdfstring{(f, g, h)}{fgh}.}
In \oursmask{} (Eq.~\eqref{eq:our-v}), the functions $f(t)$, $g(t)$, and $h(t)$ play a crucial role in enforcing the boundary conditions. While we proposed $g(t) = \cos(\frac{\pi}{2}t)$, $f(t) = \sin(\frac{\pi}{2}t)$, and $h(t) = \sin(\pi t)$ as a natural choice, other functions satisfying the boundary conditions outlined in Section~\ref{sec::method} are possible.  To assess the sensitivity of our model to these function choices, we experiment with several alternatives for $f(t), g(t), h(t)$.  Specifically, we compare the performance using:

\begin{itemize}
\item \textbf{Standard Cosine~(Default):} $g(t) = \cos(\frac{\pi}{2}t)$, $f(t) = \sin(\frac{\pi}{2}t)$, $h(t) = \sin(\pi t)$.
\item \textbf{Offset Cosine:} $g(t) = \cos(\frac{\pi}{2}t)$, $f(t) = 1 - \cos(\frac{\pi}{2}t)$, $h(t) = \sin(\pi t)$.
\item \textbf{Quadratic Polynomial:} $g(t) = 1-t^2$, $f(t) = t^2$, $h(t) = t^2(1-t^2)$.
\item \textbf{Square Root Polynomial:} $g(t) = (1-\sqrt{t})$, $f(t) = \sqrt{t}$, $h(t) = \sqrt{t}(1-\sqrt{t})$.
\item \textbf{Linear:} $g(t) = (1-t)$, $f(t) = t$, $h(t) = t(1-t)$.
\end{itemize}

We train models using each of these function sets on the ImageNet dataset and evaluate their performance using FID and IS score. The quantitative results, summarized in Figure~\ref{fig::ablation_functions_numbers}, reveal that the Standard Cosine~(Default) function choice consistently achieves the best performance. 
We provide qualitative samples generated by the these model, visualized in Appendix~\ref{sec::appendix_ablation_function}. 
The default Standard Cosine model exhibits better visual fidelity compared to those from other function choices.

\begin{table}[ht!]
\centering
\resizebox{\columnwidth}{!}{
\begin{tabular}{ccc|cccc}
\toprule
\makecell{Right\\Boundary} & \makecell{Left\\Boundary} & \makecell{2-Pass\\Model} & Euler-FID~$(\downarrow)$ & Euler-IS~$(\uparrow)$ & SDE-FID~$(\downarrow)$  & SDE-IS~$(\uparrow)$ \\
\midrule
\xmark & \xmark & \xmark & 6.87 & 130.08 & 6.79 & 141.86 \\
\checkmark & \xmark & \xmark & 6.80 & 125.44 & 6.75 & 134.37 \\
\checkmark & \checkmark & \xmark & 6.63 & 122.35 & \textbf{6.18} & 141.43 \\
\checkmark & \xmark & \checkmark & \textbf{6.32} & \textbf{137.28} & 6.38 & \textbf{149.00} \\
\bottomrule
\end{tabular}
}
\caption{
 \textbf{Boundary enforcement ablation results on ImageNet.}  We present FID, IS, Precision, and Recall scores for an ablation study comparing different boundary enforcement methods within Rectified Flow on ImageNet 256×256.  We evaluate: \rf{} (None), Right-Boundary Model (t=1 only), Double-Boundary Model (t=0 and t=1), and \oursdualpass{} (t=1, subtraction-based).  The results demonstrate that boundary enforcement consistently improves performance over \rf{}.  The \oursdualpass{} achieves the best FID and IS, indicating the effectiveness of enforcing at least the right (t=1) boundary, even with a simpler parameterization.
 }
 \vspace{-5pt}
 \vspace{-5pt}
\label{tab::ablation_boundary_enforcement}
\end{table}

\paragraph{Single- vs. Double-Boundary Enforcement.}
Next, we compare the performance of our single and double-boundary model and the subtraction-based model. This comparison isolates the impact of enforcing both boundary conditions versus primarily focusing on the $t=1$ constraint. We train all model variants using the default Standard Cosine functions under identical settings and evaluate them using FID and IS score.

The results, shown in Table~\ref{tab::ablation_boundary_enforcement}, indicate that while both \oursmask{} outperform the \rf{} baseline in terms of FID, the one with double-boundary condition achieves slightly superior performance across most metrics than its counterpart with only right boundary, particularly in FID. The \oursdualpass{} provides a significant improvement over the baseline, demonstrating the effectiveness of enforcing at least the $t=1$ boundary condition. 
We note that \oursdualpass{} introduces approximately $2\times$ increase in inference cost compared to the \oursmask{}, due to the extra forward pass for $t=1$. 

\begin{table}[ht!]
\centering
 \vspace{-5pt}
\resizebox{\columnwidth}{!}{
\begin{tabular}{l|cccc}
\toprule
Noise Scale $c$ & 0 & 1 & 2 & 5 \\
\midrule
Vanilla RF Model  & 7.18~/~128.09 & 7.64~/~132.89 & 8.26~/~135.49 & 11.76~/~126.24 \\
Ours~(Mask-based) & 6.65~/~125.71 & \textbf{6.68}~/~137.94 & \textbf{7.27}~/~141.34 & \textbf{10.68}~/~129.91 \\
Ours~(Subtraction-based) & \textbf{6.59}~/~\textbf{135.97} & 7.48~/~\textbf{142.31} & 8.04~/~\textbf{142.01} & 11.67~/~\textbf{130.74} \\
\bottomrule
\end{tabular}
}
\caption{
 \textbf{Impact of noise scale $c$ of overshooting sampler.} FID and IS scores for \rf{} and \ours{} with varying stochasity strength of Overshooting sampler ($c = 1.0, 2.0, 5.0$) on ImageNet dataset. \ours{} demonstrates more stable performance and less degradation in sample quality for varying $c$. Samples are generated using 100 steps. Numbers are shown in format FID/IS. 
 }\vspace{-5pt}
 \vspace{-5pt}
\label{tab::sde_step_ablation}
\end{table}

\subsection{Impact on Stochastic Sampling}\label{sec::impact_stochastic_sampling}

To empirically validate the impact of boundary conditions on stochastic sampling stability, we conduct experiments to compare the performance of \rf{} and \ours{} under varying degrees of stochasticity in the sampling process.  We focus on evaluating the sample quality generated by both models when using stochastic samplers with different noise scales.

We compared \rf{} and \ours{} using the Overshooting sampler~\cite{hu2024amo} with varying noise scale $c$, controlling the strength of stochasticity. We evaluated FID and IS scores across different $c$ values, maintaining consistent settings (ImageNet 256x256, 100 steps).

Table \ref{tab::sde_step_ablation} shows that \ours{} generally outperforms \rf{} across different $c$ values. Importantly, as stochasticity $c$ increases, \oursmask{} and \oursdualpass{} demonstrate better performance than \rf{}. This highlights that boundary conditions mitigate the stochastic sampling instability by fixing the score function. 
We present qualitative differences in the appendix~\ref{sec::appendix_impact_stochastic_sampling}.

\subsection{Scalability to Large-Scale Models and High-Resolution Images}
\label{sec:scalling_law}

Having demonstrated the effectiveness of boundary-aware Rectified Flow and analyzed model variants, we now assess the scalability of \ours{} to larger models and high-resolution image generation. 

Table~\ref{tab::scaling_law} shows the performance gain achieved, confirming a consistent trend with our DiT-B $256\times 256$ findings. Our subtraction-based boundary model remains universally superior to vanilla RF when scaled to DiT-L/XL and $512 \times 512$.
The mask-based boundary model also generally outperforms vanilla RF when scaling up, though IS slightly decreases for DiT-L with Euler sampler, similar to the observations on DiT-B. These results demonstrate the scalability and consistent advantages of our method. 

We show qualitative samples in Appendix~\ref{sec::appendix_scaling_vis}, which further illustrates the visual superiority of \ours{} at larger model and higher resolution.

\begin{table}[ht!]
\centering
\vspace{-5pt}
\resizebox{1.0\columnwidth}{!}{
\begin{tabular}{l|cccc}
\toprule
Method & Euler-FID~$(\downarrow)$ & Euler-IS~$(\uparrow)$ & SDE-FID~$(\downarrow)$  & SDE-IS~$(\uparrow)$ \\

\midrule
\texttt{DiT-L, $256 \times 256$} \\
Vanilla RF Model & 4.72 & 165.44 & 6.77 & 165.31 \\
Ours~(Mask-based) & 4.23 & 162.34 & 4.86 & 174.34 \\
Ours~(Substraction-based) & \textbf{3.75} & 
\textbf{187.93} & \textbf{4.52} & \textbf{191.49}\\
\hline
\texttt{DiT-XL, $256 \times 256$} \\
Vanilla RF Model & 4.04 & 165.86 & 4.36 & 171.30\\
Ours~(Mask-based) & 3.85 & 170.18 & 4.17 & 183.59\\
Ours~(Substraction-based) & \textbf{3.38} & \textbf{187.88} & \textbf{4.12} & \textbf{190.34} \\
\hline
\texttt{DiT-B, $512 \times 512$} \\
Vanilla RF Model & 16.35 & 69.92 & 16.70 & 72.93 \\
Ours~(Mask-based) & 15.21 & 70.35 & 15.69 & 77.25 \\
Ours~(Substraction-based) & \textbf{15.08} & \textbf{75.96} & \textbf{14.58} & \textbf{80.92}\\
\bottomrule
\end{tabular}
}
\caption{\textbf{Results on larger model size (DiT-L/2 and DiT-XL/2) and higher resolutions ($512 \times 512$).} Results on the larger model and higher resolutions show consistent performance improvement than vanilla Rectified Flow models. }
\vspace{-5pt}
 \vspace{-5pt}
\label{tab::scaling_law}
\end{table}

\section{Conclusion}
\label{sec::conclusion}

In this work, we identified and addressed a critical limitation in standard Rectified Flow: the violation of boundary conditions in learned velocity fields. To mitigate this, we introduced \ours{}, a novel boundary-aware parameterization that explicitly enforces these conditions through mask-based and subtraction-based variants. Extensive experiments on CIFAR-10 and ImageNet datasets demonstrate the superior performance of \ours{} over \rf{} across various quantitative metrics and qualitative assessments. Ablation studies further validated the effectiveness of our boundary-aware design. These results underscore the importance of proper boundary condition handling in Rectified Flow and establish \ours{} as a robust approach for high-quality generative modeling.  Future work could explore the application of boundary-aware parameterizations to other flow-based or diffusion generative models and investigate adaptive boundary enforcement strategies for further performance improvements and task-specific customization.   

Future work could explore extending \ours{} to larger models/higher resolutions on text-to-image/video generation task. 

\clearpage
{
    \small
    \bibliographystyle{ieeenat_fullname}
    \bibliography{main}
}

\newpage
\appendix
\onecolumn

\section{Appendix}

\subsection{Experiment Details}~\label{sec::training-details} 

\vspace{-5pt}
\paragraph{Model Configurations and Hyperparameter Settings.}
Hyperparameter configurations for CIFAR-10 and ImageNet are detailed in Table~\ref{tab:hyperparameters}. For ImageNet experiments, we employed a VAE model for initial 8×8 downsampling, converting $256 \times 256$ images to $32 \times 32$ inputs for the DiT model.  We evaluated performance metrics, including FID, IS, Precision, and Recall, on 50K generated samples.  During training, we adopted the logit normal sampling strategy as proposed in SD3~\cite{esser2024scaling}.

\begin{table*}[ht!]
\centering
\small
\resizebox{0.3\textwidth}{!}{
\begin{tabular}{lccc}
\toprule
Hyperparameters & CIFAR-10 & ImageNet \\
\midrule
Image size & $32 \times 32$ & $256 \times 256$ \\
CFG scale & - & 1.25 \\
Model &DDPM++  & DiT-B/2 \\
Batch size &512  & 768 \\
Training epoches &3072  & 350 \\
Learning rate &1e-3 & 5e-4 \\
\bottomrule
\end{tabular}
}

\caption{Hyperparameter configuration for experiments on CIFAR-10 and ImageNet dataset.}
\label{tab:hyperparameters}
\end{table*}

\vspace{-5pt}
\paragraph{Stochastic Sampling Configurations.}
We utilize the Stochastic Curved Euler Sampler (SDE Sampler in our paper)~\cite{let2025Liu} as our primary sampling method. The detailed sampling algorithm is presented in Algorithm~\ref{alg:srf}.  Additionally, for comparison, we also employ the Overshooting Sampler~\cite{hu2024amo} shown in Algorithm~\ref{alg:overshooting}.  Theoretically, the Stochastic Curved Euler Sampler is equivalent to the Overshooting Sampler with $c=1.0$ in the limit of infinitesimal step sizes, and can be viewed as a variant of DDPM adapted for Rectified Flow models~\cite{let2025Liu}.

\begin{algorithm*}[h]
\captionof{algorithm}[srf]{\atphantom\ \ Stochastic Curved Euler Sampler for Rectified Flow.}
\begin{spacing}{1.1}
\begin{algorithmic}[1] 
 \AProcedure{StochasticCurvedEulerSampler}{$v, ~\{t_i\}_{i=0}^N$}
    
    \AState{{\bf Initialize} $\tilde{\Z}_0 \sim \mathcal{N}\big(\vec{0}, ~\vec{I}\big)$}         
    \AFor{$i \in \{0, \dots, N-1\}$}         
        \AState{Calculate velocity $\vv_i = v(\tilde{\Z}_{t_i}; t_i)$} 
        \AState{Predict end points:}  
        \AState{~~~~ ~ \( \tilde{\Z}_{t_i}^{(1)} \leftarrow \tilde{\Z}_{t_i} + (1 - t_i) \vv_i \)~~~~~~ \( \tilde{\Z}_{t_i}^{(0)} \leftarrow \tilde{\Z}_{t_i} - t_i \vv_i \)
        }
        \AState{Sample random noise \( \xxi_i \sim \mathcal{N}(0, \mathbf{I}) \)} 
        \AState{Calculate noise replacement factor \( \alpha_i \leftarrow 1 - \frac{t_i (1 - t_{i+1})}{t_{i+1} (1 - t_i)} \)} \\
        \AState{Refresh end point \( \tilde{\Z}_{t_i}^{(0), \text{ref}} \leftarrow (1 - \alpha_i) \tilde{\Z}_{t_i}^{(0)} + \sqrt{1 - (1 - \alpha_i)^2} \xxi_i \)}\\
        \AState{Update sample \( \tilde{\Z}_{t_{i+1}} \leftarrow t_{i+1} \tilde{\Z}_{t_i}^{(1)} + (1 - t_{i+1}) \tilde{\Z}_{t_i}^{(0), \text{ref}} \)}
    \EndFor
    \AState{\textbf{return} $\tilde{\Z}_{t_N}$}
  \EndProcedure
\end{algorithmic}
\end{spacing}
\label{alg:srf}
\end{algorithm*}

\begin{algorithm*}[h]
\captionof{algorithm}[srf]{\atphantom\ \ Overshoot Sampling for Rectified Flow.}
\begin{spacing}{1.1}
\begin{algorithmic}[1] 

 \AProcedure{OvershootingSampler}{$v, ~\{t_i\}_{i=0}^N, ~ c \in \R^+$}
    
    \AState{{\bf Initialize} $\tilde{\Z}_0 \sim \mathcal{N}\big(\vec{0}, ~\vec{I}\big)$}         
    \AFor{$i \in \{0, \dots, N-1\}$}         
        \AState{Calculate velocity $\vv_i = v(\tilde{\Z}_{t_i}; t_i)$}          
        \AState{Overshooted ODE update:}  \\   
        \AState{~~~~ ~ $\displaystyle 
        \hat {\Z}_{o} = \tilde{\Z}_{t_i} + (\vec{o} - t_i) \circ \vv_i, 
        $~~~with~~~ $\displaystyle 
         \hspace{0pt} o= \text{min}\big( t_{i+1} + c (t_{i+1} - t_i) , ~1\big). 
         $  
        }\\ 
        \AState{Backward update by adding noise:} \\
    \AState{ $
    \displaystyle \tilde \Z_{t_{i+1}} \gets a \hat {\Z}_{o} + b \xxi_i,$ 
    ~~~~\text{where}~
    $\displaystyle\xxi_i \sim \mathcal{N} \big(\vec{0}, \vec{I}\big),$ and 
        $\displaystyle
        a = \frac{s}{o}~~~\text{and}~~~b = \sqrt{(1 - s)^2 - (a (1 - o))^2}. 
        $} \\ 
    \EndFor
    \AState{\textbf{return} $\tilde{\Z}_{t_N}$}
  \EndProcedure
\end{algorithmic}
\end{spacing}
\label{alg:overshooting}
\end{algorithm*}

\begin{figure}[ht!]
\centering
\includegraphics[width=1.0\columnwidth]{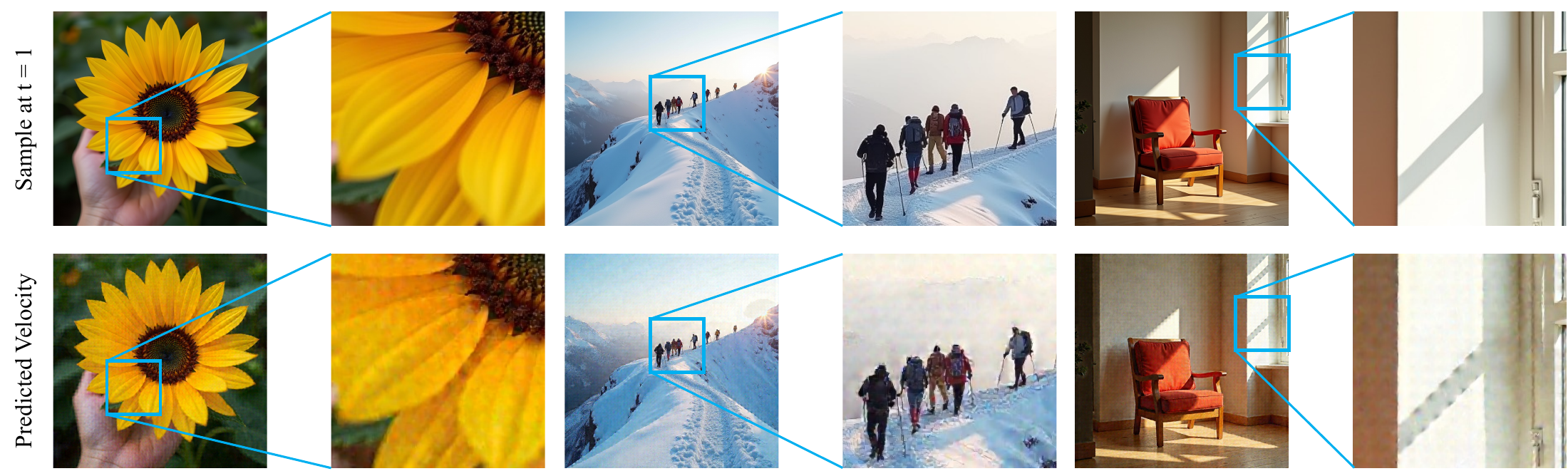}
\caption{\textbf{Boundary Condition Violation in FLUX.1-dev Model.} Visualization of the predicted velocity field at \(t=1\), demonstrating deviation from the expected data distribution and violation of the right boundary condition \(v(\mathbf{x}, t) = \mathbf{x}\).}
\label{fig:flux}
\end{figure}

\subsection{Figure~\ref{fig:teaser} Generation Details and Violation of Boundary Condition in T2I Models}~\label{sec::flux}

To generate Figure~\ref{fig:teaser} (main paper), we analyzed the velocity field predicted by the pre-trained Rectified Flow model, \textit{FLUX.1-dev}~\cite{blackforestlabs_flux_2023}.  Specifically, we extracted the model's velocity prediction at timestep $t=1$. To visualize this high-dimensional velocity field as an image, we utilized the pre-trained VAE model from the same FLUX.1-dev model. This VAE decoder was employed to map the predicted velocity vector back into the pixel space, resulting in the visual representations shown in Figure~\ref{fig:teaser}. Our analysis reveals that even this state-of-the-art text-to-image diffusion model, \textit{FLUX.1-dev}, exhibits a significant deviation from the expected boundary condition at $t=1$. This boundary violation shows a practical limitation inherent in current Rectified Flow models, as highlighted in our main paper. In Figure~\ref{fig:flux}, we show more samples generated by FLUX model that violates the boundary conditions. 

\begin{figure}[ht!]
\centering
\includegraphics[width=1.0\columnwidth]{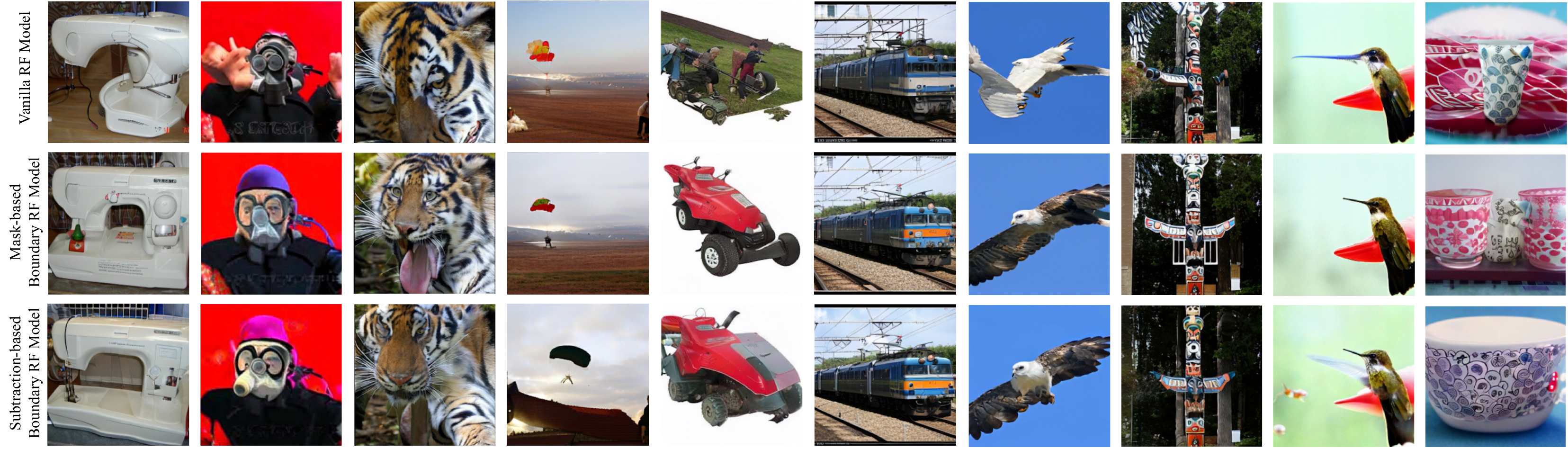}
\caption{\textbf{Qualitative comparison of image generation results on ImageNet $256 \times 256$ dataset.} We present paired examples generated by \rf{} (first row), \oursmask{} (second column) and \oursdualpass{} (last column). We use the same random seed for all models to ensure a fair visual comparison.  Our approach consistently generates images with clearer and more detailed structures, and improved visual fidelity compared to \rf{}. }
\label{fig:appendix_rf_ours_images}
\end{figure}

\subsection{Additional Qualitative Results}~\label{sec::additional-examples}
We present an expanded qualitative comparison in Figure~\ref{fig:appendix_rf_ours_images}, providing more diverse examples as an extension of Figure~\ref{fig:rf_ours_images} in the main paper.  These additional examples further show the visual advantages of \ours{} and \oursdualpass{} over the \rf{}.

\subsection{Qualitative Comparison of Boundary Functions}~\label{sec::appendix_ablation_function}

\begin{figure}[ht!]
\centering
\includegraphics[width=0.9\columnwidth]{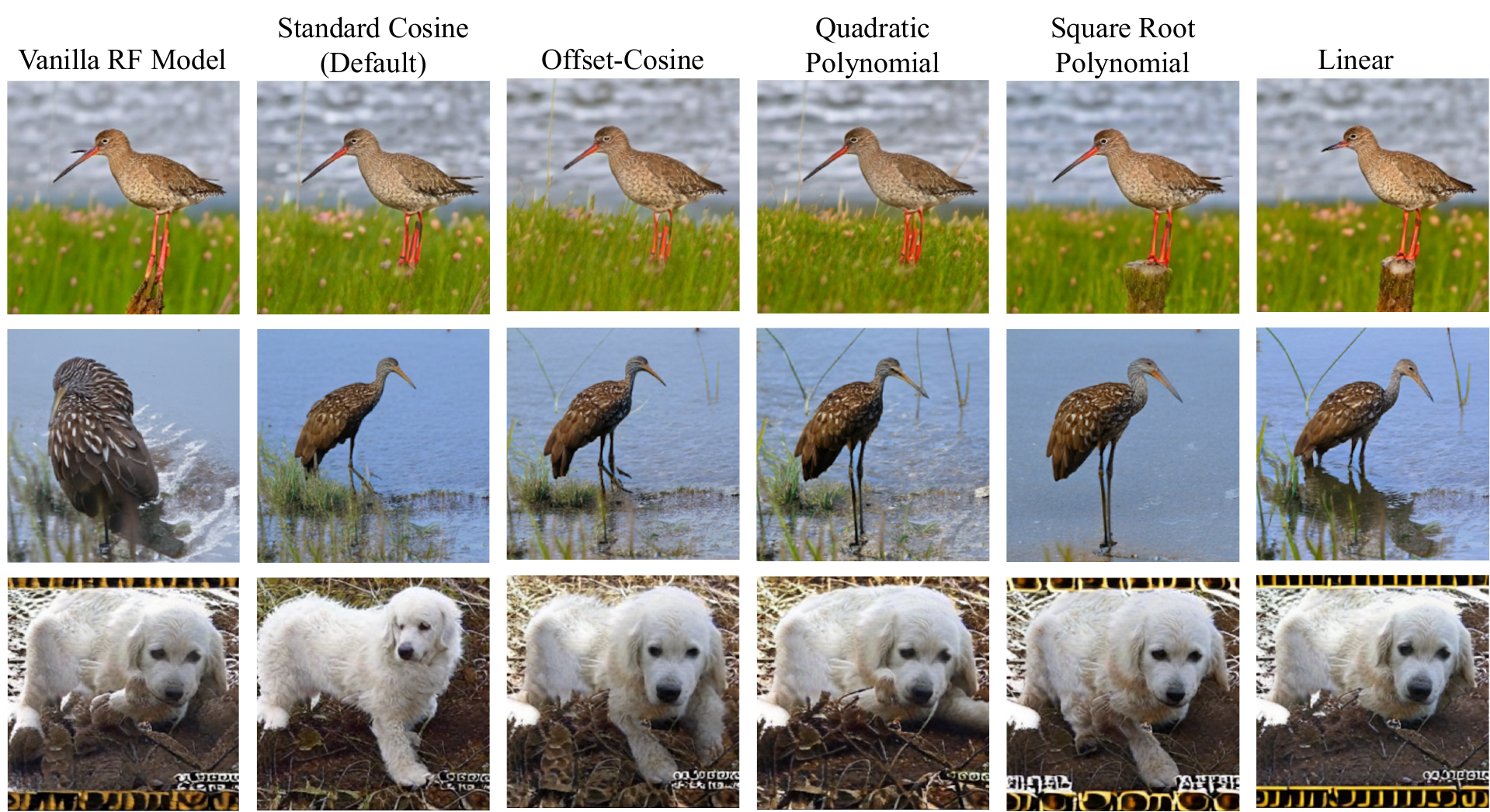}
\caption{Qualitative comparison of boundary functions.  Generated samples from double-boundary models using different choices of functions ($f(t), g(t), h(t)$).}
\label{fig::ablation_functions_samples}
\vspace{-1em}
\end{figure}

We provide qualitative samples generated by the the models that apply different boundary functions, visualized in Figure~\ref{fig::ablation_functions_samples}. 
The default Standard Cosine model exhibits better visual fidelity compared to those from other function choices.

\subsection{Qualitative Comparison of Stochastic Sampling with Overshooting Sampler.}\label{sec::appendix_impact_stochastic_sampling}

\begin{figure}[ht!]
\centering
\includegraphics[width=0.9\columnwidth]{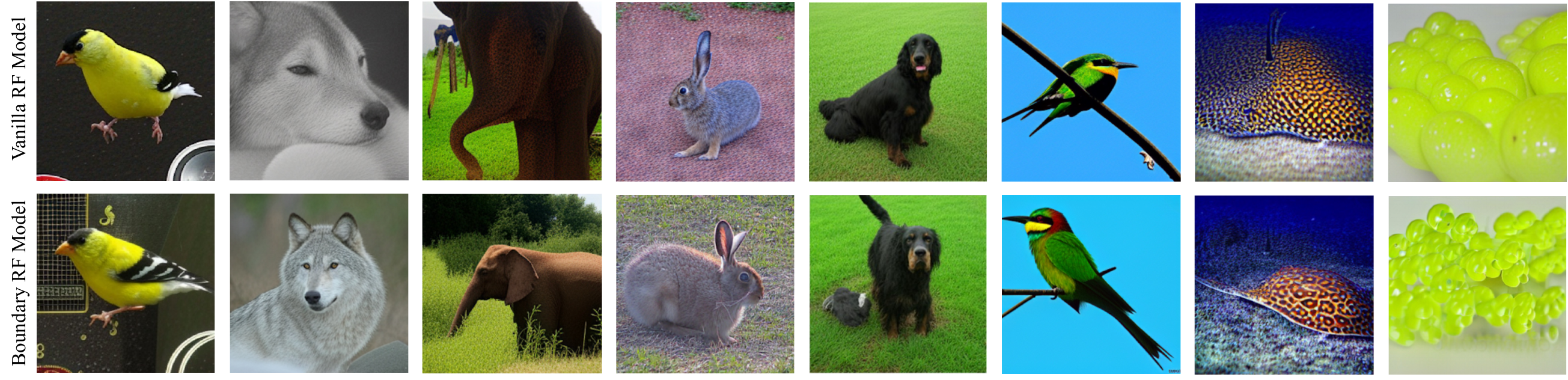}
\caption{\textbf{Qualitative comparison of stochastic sampling with Overshooting sampler.}  Qualitative samples using the overshooting sampler reveal that \ours{} effectively preserves high-frequency details, contrasting with the over-smoothed results from \rf{} under stochastic sampling.}\vspace{-5pt}
\label{fig:sde_smoothing}
\end{figure}

We present qualitative differences in Figure~\ref{fig:sde_smoothing}. We observed that over-smoothed samples from \rf{} when using Overshooting sampler, while \ours{} maintains sharper and more finegrained details.
The results illustrate the robustness of \ours{} to stochastic sampling and its superior preservation of high-frequency details compared to \rf{} under stochasticity sampling.

\subsection{Qualitative Comparison of \ours{} and Vanilla RF Model on DiT-L/2 and DiT-XL/2 Models}~\label{sec::appendix_scaling_vis}

\begin{figure}[ht!]
\centering
\includegraphics[width=0.95\columnwidth]{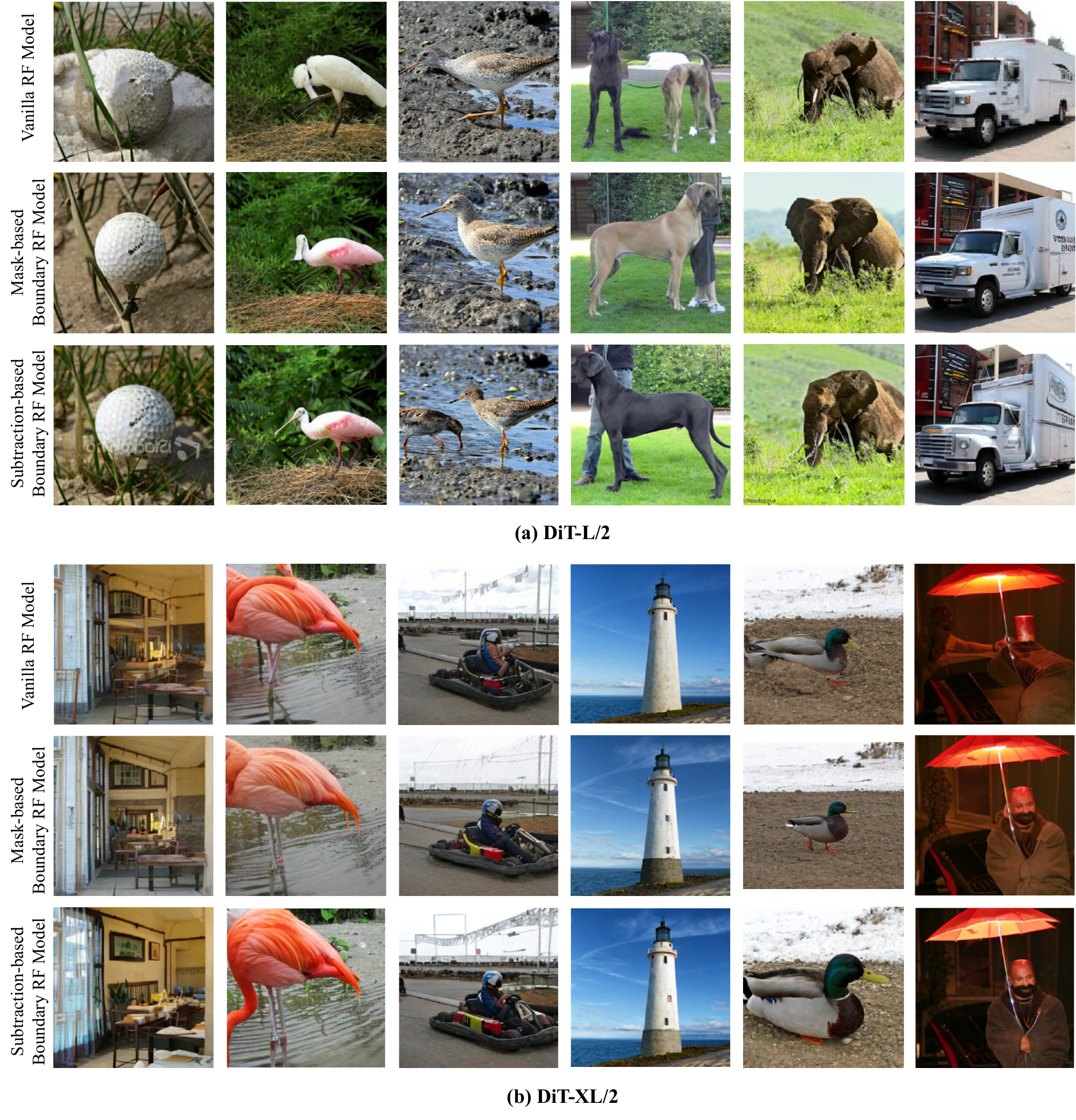}
\caption{\textbf{Qualitative comparison of image generation results on ImageNet $256 \times 256$ dataset with DiT-L/2 and DiT-XL/2 models.} We present paired examples generated by \rf{} (first row), \oursmask{} (second column) and \oursdualpass{} (last column). We use the same random seed for all models to ensure a fair visual comparison.  Our approach consistently generates images with clearer and more detailed structures, and improved visual fidelity compared to \rf{}. }
\label{fig:appendix_l_xl_dit}
\end{figure}

We present qualitative comparison in Figure~\ref{fig:appendix_l_xl_dit}, providing visual examples generated using the same random seeds. These examples show the visual advantages of \ours{} and \oursdualpass{} over the \rf{} on models with larger sizes.

\end{document}